\documentclass[a4paper,twoside]{article}

\usepackage{a4}
\usepackage{amssymb}
\usepackage{amsthm}
\usepackage{authblk}
\usepackage{url}
\usepackage{tikz}
\usepackage{amsmath}
\usepackage{float}%
\usepackage{xcolor}
\usepackage{amsmath}
\usepackage{float}
\usetikzlibrary{arrows,matrix}

\numberwithin{equation}{section}

\title{Local Approximations, Real Interpolation and Machine Learning}
\author[1]{Eric Setterqvist}
\affil[1]{Johann Radon Institute for Computational and Applied Mathematics (RICAM), Austrian Academy of Sciences, Linz, Austria, email: eric.setterqvist@ricam.oeaw.ac.at}
\author[2]{Natan Kruglyak}
\affil[2]{Department of Mathematics, Link\"oping University, Sweden, e-mail: natan.kruglyak@liu.se}
\author[3]{Robert Forchheimer}
\affil[3]{Department of Electrical Engineering, Link\"oping University, Sweden, and RISE Research Institutes of Sweden AB, e-mail: robert.forchheimer@liu.se}

\begin{document}
\maketitle

\begin{abstract}

 We suggest a novel classification algorithm that is based on local
approximations and explain its connections with Artificial Neural Networks
(ANNs) and Nearest Neighbour classifiers. We illustrate it on the datasets MNIST
and EMNIST of images of handwritten digits. We use the dataset MNIST to find
parameters of our algorithm and apply it with these parameters to the challenging
EMNIST dataset. It is demonstrated that the algorithm misclassifies 0.42\% of the images
of EMNIST and therefore significantly outperforms predictions by humans and
shallow artificial neural networks (ANNs with few hidden layers) that both
have more than 1.3\% of errors.

\end{abstract}

\section{Introduction\label{intro}}
The Nearest Neighbour (NN) classifier is one of the most basic classification
methods and was described and analysed for the first time in 1951 \cite{FixHod51, SilJon89}. The NN classifier compares an unknown object with a set of labelled
object and the label of the closest object (based on some distance metric) is selected as the classifier result \cite[pp.~124--127]{Bis06}.

Today, state of the art classifiers are based on Artificial Neural Networks (ANNs). However, in spite of their remarkable success ANNs work as
black boxes. The non-existence of a mathematical theory of
neural networks makes it difficult to understand when and why ANNs
works. It is unclear how to interpret, and thereby prevent, their failures. So, it would be interesting to develop a classifier that is based on
rigorous mathematics and that provides results on par with ANN-based methods. To
construct such a classifier we need to learn from the ANNs, in particular to
incorporate the properties that make ANNs so powerful when it comes to their
predictive performance.

In this paper we will investigate analogues between neural networks and local approximations and let these guide us in the construction of a novel classification algorithm. The resulting classifier is firmly rooted in approximation theory and real interpolation. As our approach could be considered as a windowed version of the Nearest Neighbour classifier we will call the classifier Windowed Nearest
Neighbour, or WNN for short. While the present work concerns the mathematical aspects of the WNN classifier, we also include for the reader's convenience experimental results of \cite{SetKruFor22}. These results show that the WNN classifier gives results not far from ANNs on MNIST and EMNIST datasets of images of handwritten digits.

\section{Mathematical roots of the algorithm}

\subsection {Local approximation and a result in real interpolation} \label{sec:approximation}

We recall that the modern theory of local approximations was developed in the
1970s by Yu. Brudnyi (see \cite{brudnyi}) who used them to describe spaces of differentiable
functions by their local approximations by polynomials of fixed degree. Let us
formulate one result from Brudnyi's theory.

Let $f\in L^{p}(Q_{0}),1<p<\infty$ and $Q_{0}$ be a cube in
$\mathbb{R}^{n},$ here and everywhere below we suppose that cube faces are parallel to the coordinate hyperplanes. Let $Q$ be a cube in $\mathbb{R}^{n}$ with center in $Q_{0}$.
Then the quantity%
\begin{equation}
E_{k}(f,Q)_{p}=\inf_{P}(\int_{Q\cap Q_{0}}\left\vert f(x)-P(x)\right\vert
^{p}dx)^{1/p},\label{1}%
\end{equation}
where infimum is taken over all polynomials $P$ of degree strictly less than
$k$ is called a local approximation of function $f$. Let us consider the well-known
in approximation theory $k$-modulus of continuity%
\[
\omega_{k}(f,t)_{p}=\sup_{\left\vert h\right\vert <t}\left\Vert \sum_{j=0}%
^{k}(-1)^{k-j}\frac{k!}{j!(k-j)!}f(x+jh)\right\Vert _{L^{p}},
\]
where sup is taking over all $h\in\mathbb{R}^{n}$ such that $\left\vert
h\right\vert <t$ and $x,x+h,...,x+kh\in Q_{0}$.

In \cite{brudnyi} Brudnyi showed that with constants of equivalence independent of $f$ and $t>0$
we have%
\[
\omega_{k}(f,t)_{p}\approx\sup_{\left\{  Q_{i}\right\}  }\left(  \sum_{Q_{i}%
}(E_{k}(f,Q_{i})_{p})^{p}\right)  ^{1/p},
\]
where sup is taken over all finite families $\left\{  Q_{i}\right\}  $ of
cubes $Q_{i}$ with centers in $Q_{0}$ with side length equal to $t$ and disjoint
interiors. It was indicated by Peetre \cite{Pee68} that the modulus of continuity $\omega
_{k}(f,t)_{p}$ is deeply connected with the $K$-functional of real
interpolation. Brudnyi showed later in \cite{brudnyi} that for the couple $(L^{p},\dot{W}%
_{p}^{k})$ on the cube $Q_{0}$, where $\dot{W}_{p}^{k}$ is a homogenous Sobolev
space defined by finiteness of the quasinorm
\[
\left\Vert f\right\Vert _{\dot{W}_{p}^{k}}=\sup_{k_{1}+...+k_{n}=k}\left\Vert
\frac{\partial^{k_{1}}}{\partial^{k_{1}}x_{1}}...\frac{\partial^{k_{n}}%
}{\partial^{k_{n}}x_{n}}f\right\Vert _{L^{p}},
\]
the $K$-functional%
\[
K(t,f,L^{p},\dot{W}_{p}^{k})=\inf_{g\in\dot{W}_{p}^{k}}(\left\Vert
f-g\right\Vert _{L^{p}}+t\left\Vert g\right\Vert _{\dot{W}_{p}^{k}}),\text{
\ }t>0,
\]
is equivalent to the modulus of continuity
\[
K(t^{k},f,L^{p},\dot{W}_{p}^{k})\approx\omega_{k}(f,t)_{p}%
\]
with constants of equivalence independent of $f$ and $t$. So, the
$K$-functional of the couple $(L^{p},\dot{W}_{p}^{k})$ can be described in
terms of local approximations
\[
K(t^{k},f,L^{p},\dot{W}_{p}^{k})\approx\sup_{\left\{  Q_{i}\right\}  }\left(
\sum_{Q_{i}}(E_{k}(f,Q_{i})_{p})^{p}\right)  ^{1/p},
\]
where sup is taken over all finite families $\left\{  Q_{i}\right\}  $ of
cubes $Q_{i}$ with centers in $Q_{0}$ with side length equal to $t$ and disjoint interiors.

Later, see \cite[Thm.~9.2]{KisKru13}, expressions (in terms of local approximations)
for the $K$-functionals of the couples $(L^{p_{0}},\dot{W}_{p_{1}}^{k})$ were found. In
particular, a different formula for the $K$-functional of the couple
$(L^{p},\dot{W}_{p}^{k})$ for functions on $\mathbb{R}^{n}$ was found. To formulate it
let us split $\mathbb{R}^{n}$ on cubes $Q_{i}$ with side length equal to $t$ and
consider family of cubes $\left\{  K_{i}\right\}  $, where cube $K_{i}$ has
the same center as cube $Q_{i}$ and side length $\frac{3}{2}t$ (note that
neighbour cubes $K_{i}$ and $K_{j}$ intersect). Then%
\begin{equation}
K(t^{k},f,L^{p},\dot{W}_{p}^{k})\approx\left(  \sum_{K_{i}}(E_{k}(f,K_{i}%
)_{p})^{p}\right)  ^{1/p}=\left(  \sum_{K_{i}}\inf_{\deg(P)<k}(\int_{K_{i}%
}\left\vert f(x)-P(x)\right\vert ^{p}dx)\right)^{1/p}.\label{2}%
\end{equation}
That is, we do not need to take supremum over all finite families $\left\{
Q_{i}\right\}  $ with side length equal to $t^{\text{ }}$and disjoint interiors.

The right hand side of (\ref{2}) suggests the following classification algorithm.

\subsection {A Classification Algorithm based on Local Approximations}

Let $A_{1},...,A_{M}$ be some sets in $L^{p}$. Then for any cube $Q$ with
center in $Q_{0}$  we can define
$M$ local approximations according to%
\[
E_{A_{m}}(f,Q)_{p}=\inf_{g\in A_{m}}(\int_{Q\cap Q_{0}}\left\vert
f(x)-g(x)\right\vert ^{p}dx)^{1/p},\text{ \ }m=1,...,M.
\]
Note that above, in subsection \ref{sec:approximation}, we consider the case when $M=1$ and $A_{1}$ is the set of
polynomials of degree strictly less than $k$. Suppose also that some family
$\left\{  W_{i}\right\}  $ of cubes ("windows") with centers in $Q_{0}$ and equal side length are given. Then the right hand side of (\ref{2}) suggests to
consider $M$ "distances"%
\[
Dist(f,A_{m})=\left(  \sum_{W_{i}}(E_{A_{m}}(f,W_{i})_{p})^{p}\right)
^{1/p},\text{ \ }m=1,...,M.
\]
Our classification algorithm classifies $f$ as from class $m$ if%
\[
Dist(f,A_{m})=\min_{j=1,...,M}Dist(f,A_{j}).
\]
In the case when minimum is attained for several indices $1\leq j_{1}%
<...<j_{n}\leq M$ we will (arbitrarily) classify $f$ as an image of the class
$A_{j_{1}}$. We would like to note that such a situation did not occur in our experiments.

\subsection {Connections to Nearest Neighbour classifier and Artificial Neural Networks}

Suppose that we have several classes of labeled images $A_{1},...,A_{M}$ and
$B$ is an image that we need to classify. Note, that grey images can be
considered as a function  defined on the
set of discrete points (pixels) in $\mathbb{R}^{2}$ (for colour images we need to consider three functions). We will suppose that
pixels are all points with integer coordinates and functions that corresponds
to images are equal to zero for pixels outside the screen, i.e. outside some
fixed cube that we denote by $Q_{0}$. Note that for colour images number of pixels on each window will be three times more. 
%NOTE THAT FOR COLOUR IMAGES NUMBER OF PIXELS ON EACH WINDOW WILL BE THREE TIMES MORE THAN FOR THE GREY IMAGES. 

In the NN algorithm we calculate distances from $B$ to each class $A_{j}$, 
$j=1,...,M$, and classify $B$ as an element from the class $A_{i}$ if the distance from
$B$ to $A_{i}$ is the smallest. So, the algorithm in subsection 2.2
coincides with the NN classification in the case when the set of cubes (windows)
$\left\{  W_{i}\right\}  $ consists of just one window $W=Q_{0}$. Our classification algorithm based on local approximations can therefore be considered as a windowed version of the NN classifier.

Connections with ANN classifiers are more complicated. We first note that standard feedforward neural networks with ReLU activation
function can be considered as consecutively applying the following two
parts. The first part $F_{1}$ is nonlinear and transform image $B$ that we
need to classify to some space $\mathbb{R}^{N}$ while the second part $F_{2}$ is
linear and transform $\mathbb{R}^{N}$ to $\mathbb{R}^{M}$ where $M$ is the number
of classes. Then $B$ is classified as an element of class $A_{j}$ if
coordinate $j$ is maximal in $F_{2}F_{1}(B)$. Moreover, it is possible to
prove that the nonlinear part $F_{1}$ can be seen as several (sometimes more
than hundred) consecutively applied convolution transformations $T_{k}$ with
ReLU activation function%
\[
F_{1}(B)=T_{K}(T_{K-1}(...(T_{1}(B))...))
\]
where by convolution transformation with ReLU activation function we mean the
following transformation. Let $\left\{  W_{i}\right\}
_{i=1,...,I}$ be some set of windows with equal size, i.e. each window contains the same number of pixels which we denote by $n$,
and let $L:\mathbb{R}^{n}\rightarrow
\mathbb{R}^{N}$ be some linear map with $L(x)_{+}=\max(L(x),0)$. Then by
convolution transform with ReLU activation function we mean the nonlinear
transform
\[
T(B)=(L_{+}(B_{W_{1}}),...,L_{+}(B_{W_{I}})),
\]
where $B_{W}$ is the restriction of image $B$ to the pixels from window $W$. The name
"convolution" corresponds to the property that on all windows the transformations
$L_{+}$ are the same.

Comparing with our classification algorithm based on local approximations, instead of the nonlinear mapping $F_{1}$ we consider on each window $W_{i}$ the quantity 
\begin{equation*}
((E_{A_{1}}(f,W_{i})_{p})^{p},\dots,(E_{A_{M}}(f,W_{i})_{p})^{p})
\end{equation*}
which plays the role of $L_{+}(B_{W_{i}})$. Next, the counterpart of the linear mapping $F_{2}$ is the summation over all windows for each class $A_{m}$ according to
\[
Dist(f,A_{m})^{p}=\sum_{W_{i}}(E_{A_{m}}(f,W_{i})_{p})^{p}.%
\]
Note further that, similarly to $L_{+}$, the formulas for local approximations $E_{A_{m}}(f,\cdot)_{p}$ are the same for different windows and also use restrictions of $f$ on used windows.
%the role of $L_{+}(B_{W})$ is played by $((E_{A_{i}}(f,W)_{p})^{p},...,(E_{A_{m}}(f,W)_{p})^{p})$  instead of nonlinear part $F_{1}$ we considered the set of windows of the%same size and on each window $W$ the role of $L_{+}(B_{W})$ is played by
%$((E_{A_{i}}(f,W)_{p})^{p},...,(E_{A_{m}}(f,W)_{p})^{p})$ we used instead of the second linear part $F_{2}$ the summation
%\[
%Dist(f,A_{i})^{p}=\sum_{W_{i}}(E_{A_{i}}(f,W_{i})_{p})^{p}%
%\]
%and instead of nonlinear part $F_{1}$ we considered the set of cubes of the
%same size and on each window $W$ the role of $L_{+}(B_{W})$ is played by
%$((E_{A_{i}}(f,W)_{p})^{p},...,(E_{A_{m}}(f,W)_{p})^{p})$. Note that similarly to convolution, the transform formulas for local approximation are the same for different cubes and also use restrictions of $f$ on used cubes.

\section{MNIST and EMNIST datasets}

\begin{figure}
	[h]
	\begin{center}
		\includegraphics[
		height=2.7224in,
		width=3.6253in
		]%
		{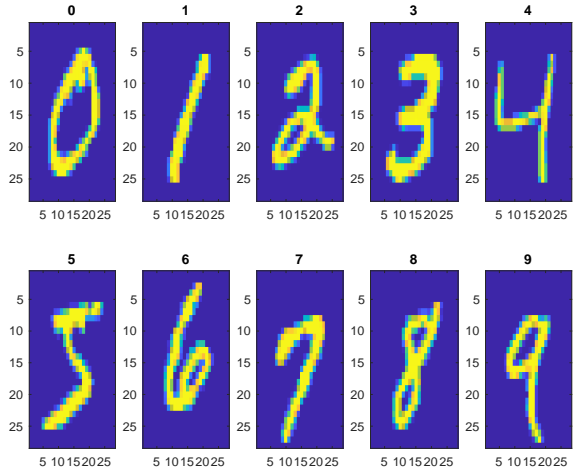}%
	\end{center}
	\caption{10 random MNIST test images. Above each image is its label.}
	\label{fig:10_test_images_mnist}
\end{figure}

We will illustrate our algorithm on MNIST \cite{mnist} and EMNIST \cite{CohAfsTapSch17} datasets. The datasets MNIST
and EMNIST contain images of handwritten digits obtained by using different
preprocessing from some parts of the NIST Special Database 19. All images in
MNIST and EMNIST datasets are greyscale images of size $28\times 28$, i.e.
they consist of 784 pixels and on each pixel the image takes an integer value
between 0 and 255. The standard MNIST dataset contains 60000 images of
handwritten digits which are usually used for training and another 10000
images of handwritten digits which typically are assigned for testing. Note
that the MNIST dataset contains images that are not too difficult for
classification, see Figure \ref{fig:10_test_images_mnist}, and the human error rate on it is reported to 0.2\% \cite{LeCJacBotCorDenDruGuyMueSacSimVap95}.

For the EMNIST dataset, we will focus on the EMNIST Digits subset containing images of handwritten digits with 4000 test images and 24000 training images for each digit $0,1,...,9$. When writing EMNIST below, we refer to this subset. 

\begin{figure}
	[h]
	\begin{center}
		\includegraphics[
		height=2.7224in,
		width=3.6253in
		]%
		{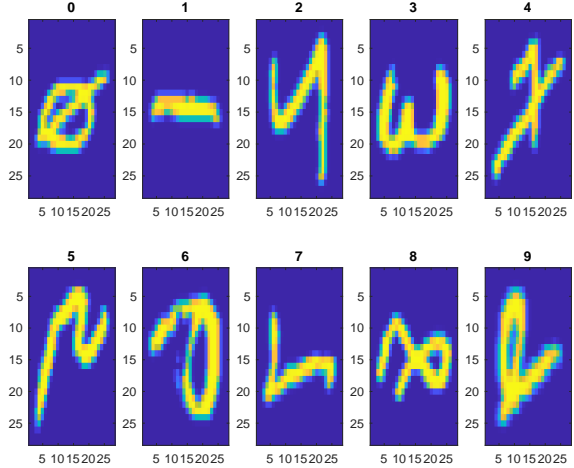}%
	\end{center}
	\caption{10 random EMNIST test images. Above each image is its label.}
	\label{fig:10_test_images_emnist}
\end{figure}

Contrary to MNIST, the images in EMNIST look more like various types of shapes than images of handwritten digits, see e.g. Figure \ref{fig:10_test_images_emnist}.
%For the construction of EMNIST, the NIST dataset was preprocessed. Probably, this is the reason that images in EMNIST look more like
%various types of shapes than images of handwritten digits, see  Figure 2. 
%\begin{figure}
%	\begin{center}
%	\includegraphics[width=0.7\columnwidth]{P1.eps}
%	\end{center}
%	\caption{Examples of images of digits 0, 1, and 2 in EMNIST. Above each image is the label.}
%	\label{fig:images_emnist}
%\end{figure}
Since we did not find any result regarding human error rate on EMNIST, we estimated our own error rate and it turned out to be more than 1.5\%. We also tried shallow neural networks (with a few hidden layers) and noticed that the error rate was more than 1.3\% for these networks. So this dataset is much more difficult for classification.

%It appears that even after a few days of training ourselves, our error rate is more than 1.5\%. We also tried shallow neural networks (with a few hidden layers) and noticed that the error rate was more than 1.3\% for these networks. Since we did not find any result regarding human error rate on this dataset, we estimated our own error rate and it turned out to be more than 1.5\%. So this dataset is much more difficult for classification.

We will also use MNIST Balanced dataset that is constructed in the following
way. Let us enumerate the images in the MNIST dataset for each digit in the order
that they appear starting with the training set and then the test set. Then,
for example, for digit 0 we will have in total 5923 training images enumerated
1 - 5923 and 980 test images enumerated 5924 - 6903. Note that there are
different number of images in the training and test sets for different digits.
To make this dataset more similar to EMNIST we will change the sets of
training and test sets in the standard MNIST dataset. More precisely, the training
set will consist of the first 6000 images for each digit and the test set will
consist of all remaining images. This new dataset we will call MNIST Balanced
dataset. Note that this dataset is uniform in size of training sets with
respect to the different digits.

The recent survey \cite{BalSaeIsa19} provides a detailed overview of image classification on MNIST and EMNIST, covering both traditional methods and ANNs, where further references can be found.

\section{WNN algorithm}

\subsection{ An algorithm.}

As we wrote above all images in MNIST and EMNIST are greyscale images of size
28 $\times\ $28, i.e. they consist of 784 pixels and on each pixel the image
takes an integer value between 0 and 255. For each of the 784 pixels, we will
consider a square `window' $W$ centered at the pixel with side length S where
S is an odd positive integer. Each window $W$ therefore consists of $S^{2}$
pixels. For simplicity of notation, we will from now on refer to S as the size
of the window. Note that for some pixels, the window $W$ will extend beyond
the image boundaries. We will set the values of the pixels in $W$ which falls
outside the screen equal to zero.

Denote by $A_{train}^{i}$, $i=1,\dots,10$, the class of training images that
corresponds to digit $i-1$. By $B_{test}^{i}$ , we denote the corresponding
class of test images for digit $i-1$. Let $B$ be some test image. We will
calculate distances between $B$ and $A_{train}^{i}$ on the window $W$
according to%
\[
dist_{W}(B,A_{train}^{i})=\min_{A\in A_{train}^{i}}(\sum_{y\in W}%
(B(y)-A(y))^{2})^{1/2},
\]
so $dist_{W}(B,A_{train}^{i})$ is a discrete analog of local approximation in
metric of $L^{2}$ if instead of polynomials $P$ of degree strictly less than
$k$ we will consider the set $A_{train}^{i}$.

Next, for each class $A_{train}^{i}$ we take into account the distances on all windows
and define%
\[
Dist(B,A_{train}^{i})=(\sum_{W}dist_{W}(B,A_{train}^{i})^{2})^{1/2}.
\]
Note that this formula is a discrete analog of the formula (\ref{2}).

Our algorithm classifies image $B$ as an image from the class $B_{test}^{i}$
if
\[
Dist(B,A_{train}^{i})=\min_{j=1,...,10}Dist(B,A_{train}^{j})\text{.}
\]

In the case when minimum is attained for several indices $1\leq i_{1}%
<...<i_{n}\leq10$ we will (arbitrarily) classify $B$ as an image of the class
$B_{test}^{i_{1}}$. It can be noted that such a situation has not occurred in
our investigations.

\subsection{ Size of used windows}

 To apply the algorithm we need to know the size $S$ of used
windows. To find it we did experiments with the MNIST Balanced dataset. 

The next table ({Table~\ref{err-basic}}) shows how many classification errors WNN produces for different window sizes $S$ (indicated by WNN$S$, for example by WNN11 we will mean the case when $S=11$ ). By 'NN' we denote the case when $S=55$, i.e. we actually have only one window and our classification algorithm coincides with the usual Nearest Neighbour algorithm. We see that WNN with $S=11$ is the best and has 106 errors from 10000 test images, i.e. the error rate is 1.06\%. From this table we also see that the NN algorithm has 266 errors, i.e. a much higher error rate than the WNN11 algorithm.

\begin{table}[H]
	\centering
	\resizebox{1.2\textwidth}{!}{%
		\begin{tabular}{ |c|c|c|c|c|c|c|c|c|c|c|c|c| } 
			\hline
			Digit & NN & WNN3 & WNN5 & WNN7 &WNN9 & WNN11 & WNN13 & WNN15 & WNN17 & WNN19 & WNN21 & WNN23\\ 
			\hline
			0 & 7 & 17 & 7 & 4 & 4 & 5 & 5 & 5 & 5 & 5 & 5 & 5\\ 
			1 & 10 & 123 & 35 & 14 & 6 & 5 & 5 & 7 & 6 & 7 & 7 & 7\\
			2 & 37 & 28 & 13 & 8 & 8 & 7 & 9 & 10 & 12 & 14 & 14 & 16\\
			3 & 48 & 42 & 16 & 16 & 16 & 14 & 12 & 13 & 18 & 19 & 19 & 21\\
			4 & 28 & 14 & 7 & 6 & 5 & 6 & 8 & 8 & 8 & 9 & 12 & 14\\
			5 & 2 & 5 & 1 & 1 & 2 & 2 & 2 & 2 & 2 & 2 & 2 & 2\\
			6 & 12 & 28 & 21 & 11 & 10 & 8 & 7 & 7 & 7 & 7 & 8 & 8\\
			7 & 43 & 76 & 38 & 28 & 19 & 18 & 16 & 20 & 22 & 25 & 26 & 27\\
			8 & 41 & 20 & 9 & 7 & 11 & 12 & 13 & 15 & 13 & 13 & 15 & 15\\
			9 & 38 & 54 & 38 & 31 & 29 & 29 & 30 & 27 & 28 & 29 & 31 & 33\\
			\hline
			Total & 266 & 407 & 185 & 126 & 110 & 106 & 107 & 114 & 121 & 130 & 139 & 148\\
			\hline
		\end{tabular}%
	}
	\caption{\label{err-basic}Errors for different window sizes on MNIST Balanced.}
\end{table}

\section{Experiments on MNIST}\label{wnn-extended}

It is expected that a larger training set will improve the performance of WNN. For this reason, we first extended the training set of MNIST Balanced (denoted `Set 0') artificially by spatially shifting each image not more than one pixel in horizontal, vertical or in both directions at the same time. This gives eight new images for each original training image and in total $60000\cdot 9=540000$ training images. We refer to this set as `Set 1'. In a second step, each training image of Set 1 was rotated $\pm 5,\pm 25$ degrees generating an additional $2.16$ million training images. This set of $2.7$ million training images is denoted `Set 2'. Next, note that the digits are contained in the center $20\times 20$ pixels \cite{mnist} of the image. For each image of Set 1 we then generated four new images by compressing/expanding the width or the height of this center part to $18$ and $22$ pixels. These compressed and expanded images together with Set 1 gives `Set 3' with in total $2.7$ million images. Finally, the images of Sets 2 and 3 together constitute `Set 4' which accordingly contains $4.86$ million unique images. In Table \ref{err_trainingsets}, we give the resulting number of errors of WNN11 for the different training sets.
\begin{table}[h]
	\centering
	%\resizebox{1.2\textwidth}{!}{%
		\begin{tabular}{ |c|c|c|c|c|c| } 
			\hline
			 Training set & Set 0 & Set 1 & Set 2 & Set 3 & Set 4\\ 
			\hline
			No. of training images & 60000 & 540000 & 2700000 & 2700000 & 4860000\\
			\hline
			No. of test images & 10000 & 10000 & 10000 & 10000 & 10000\\
			\hline
			No. of errors on test images & 106 & 62 & 49 & 49 & 41\\ 
			\hline
			Error rate & 1.06\% & 0.62\% & 0.49\% & 0.49\% & 0.41\% \\
			\hline
		\end{tabular}%
	%}
	\caption{\label{err_trainingsets}Errors for WNN11 on MNIST Balanced when using different extensions of the training set.}
	
\end{table}
We note that Set 4 gives the lowest error rate with 0.41\%. Further expansions of the training set might give better results but we chose in this study to restrict ourselves to a few straightforward alternatives. When comparing with previously published work on NN-based methods, recall that the training and test sets considered in this study are not the standard ones of MNIST. However, when applying the WNN algorithm (with extension of the training set as above) on the MNIST standard set, we obtain an error rate of 0.48\% which is lower than the best published result of 0.52\% \cite{KeyDesGolNey07} that we are aware of.
%We note that the human error rate on MNIST is measured to around 0.2\% \cite{LeCJacBotCorDenDruGuyMueSacSimVap95}.

\section{Experiments on EMINST}
\label{emnist}

Applying the WNN algorithm on EMNIST, window size 11 will be used based upon previous investigations on MNIST. The original training set of 240000 images is denoted by `Set 0' and the extension of this set by spatially shifting not more than one pixel in horizontal, vertical or in both directions at the same time is referred to as `Set 1'. Adding rotations of $\pm 5,\pm 25$ degrees of each image to this set, a set denoted `Set 2' is constructed. Finally, we consider an extension of Set 0 in terms of a spatial shift of maximum two pixels in horizontal, vertical or in both directions at the same time (denoted `Set 3') and, as in the previous case, then extend Set 3 to include rotated images by $\pm 5,\pm 25$ degrees (denoted `Set 4'). These extensions are important. Indeed, the NN algorithm on EMNIST using the training images of Set 0 gives 625 errors (an error rate of 1.56\%) and 385 errors (an error rate of 0.96\%) using Set 4. The classification results of WNN using the different training sets are given in Table \ref{emnist_trainingsets}.
\begin{table}[h]
	\centering
	%\resizebox{1.2\textwidth}{!}{%
	\begin{tabular}{ |c|c|c|c|c|c| } 
		\hline
		Training set & Set 0 & Set 1 & Set 2 & Set 3 & Set 4\\ 
		\hline
		No. of training images & 240000 & 2160000 & 10800000 & 6000000 & 30000000\\
		\hline
		No. of test images & 40000 & 40000 & 40000 & 40000 & 40000\\
		\hline
		No. of errors on test images & 303 & 195 & 190 & 177 & 168\\
		\hline
		Error rate &  0.76\% & 0.49\% & 0.48\% & 0.44\% & 0.42\%\\
		\hline
	\end{tabular}%
	%}
	\caption{\label{emnist_trainingsets}Errors for WNN11 on EMNIST when using different extensions of the training set.}
\end{table}
For comparison, we did experiments with one third of the training images (8000 images per digit) and did extension as for Set 4. The resulting number of errors for the WNN algorithm became 202 giving an error rate of 0.5\%.

The best published result of a traditional classification method applied to EMNIST that we are aware of reports an error rate of 2.26\% \cite{GhaIngSon18}. Therefore, even without extensions of the training set the result of WNN seems to be state of the art. Further, we obtain error rates which are much lower than the human error rate of 1.5\%.

We would like to finish this section with a description of a more sophisticated version of the WNN algorithm. By using this version we can reduce the number of errors from 168 to 129 (an error rate of 0.32\%) which is on par with the best neural network result (see \cite{JayJayJayRajSenRod19}).
 
Let us briefly describe this algorithm. Let $A$ be a training image from Set 0. Denote by $A_{ext}$ the set which consists of $A$ and all its extensions as described above (so $A_{ext}$ consists of 125 images). Then we define the distance $d$ from the test image $B$ to $A$ as
\begin{equation} 
d(B,A)=\Big(\sum_{W}(d_{W}(B,A_{ext}))^2\Big)^{1/2},
\end{equation}
where $d_W(B,A_{ext})$ is the distance on window W from $B$ to $A_{ext}$ given by
\begin{equation}
d_W(B,A_{ext})=\min_{X\in A_{ext}}\Big(\sum_{w\in W}(B(w)-X(w))^{2}\Big)^{1/2}.
\end{equation}
Next, the distance $D$ from $B$ to the training class $A^{i}_{train}$ from Set 0 (containing 24000 images for each digit $i-1$), $i=1,\dots,10$, is defined as
\begin{equation}  
D(B,A_{train}^{i})=\min_{A\in A^{i}_{train}}d(B,A).
\end{equation}
We classify $B$ as an image from the class $B_{test}^{i}$ if
\begin{equation*}
D(B,A_{train}^{i})=\min_{j\in\left\{1,\dots,10\right\}}D(B,A_{train}^{j}).
\end{equation*}
This distance algorithm (denoted DWNN) gives 148 errors on EMNIST. Now for each test image $B$ we do predictions by WNN and DWNN. Usually these two predictions coincide, if they are different then we classify $B$ according to the NN classifier using the corresponding two training classes from Set 4 (containing $3\times 10^6$ images each). The resulting algorithm gives 129 errors on EMNIST.
%See Table 3 in the survey \cite{BalSaeIsa19} for further results on EMNIST and NIST Special Database 19.

	\section{An Algorithm for Decreasing the Number of Used Windows}
	It is of importance to decrease the computational cost of the WNN algorithm while maintaining its predictive performance. In this section we discuss one possible approach to this problem.
	%IT IS VERY IMPORTANT TO DECREASE  CALCULATIONS WITHOUT INCREASING NUMBER OF ERRORS. IN THIS SECTION WE DISCUSS ONE APPROACH TO THIS PROBLEM.
	
	Denote by WNN$_{W_1,...,W_K}$ the WNN algorithm where the windows
	$W_{1},...,W_{K}$ are excluded, i.e. instead of $Dist(B,A_{train}^i)$ we use	
	\[
	Dist_{W_1,\dots,W_K}(B,A_{train}^{i})=\left(\sum dist_{W}(B,A_{train}^{i})^{2}\right)
	^{1/2},\,i=1,\dots,10
	\]
	where we sum over all windows $W$ except $W_{1},...,W_{K}$.
	
	We determine $W_{1},...,W_{K}$ in an iterative way according to the following. First, we calculate for each window $W$ the number of errors for the WNN$_{W}$ algorithm. This number is denoted NE$_{W}$ (number of
	errors when window $W$ is excluded). We then consider the set of windows $W$ with smallest number of errors NE$_{W}$. Usually this set contains many windows. For every window in the set, consider
	\begin{equation*}
	GAP_W=\sum_{i=1}^{10}\sum_{B\in B_{test}^{i}}(Dist_{W}(B,A_{train}^{i})-\min_{j\in\left\{1,...,10\right\}}Dist_{W}(B,A_{train}^{j}))
	\end{equation*}
	and exclude the window $W$ for which GAP$_{W}$ is maximal. To explain the idea behind this algorithm note that if the test image $B$
	is in class $B_{test}^i$ and is predicted correctly by WNN then
	\[
	Dist_{W}(B,A_{train}^{i})-\min_{j=1,...,10}Dist_{W}(B,A_{train}^{j})=0
	\]
	and if $B$ is not predicted correctly then
	\[
	Dist_{W}(B,A_{train}^{i})-\min_{j=1,...,10}Dist_{W}(B,A_{train}^{j})>0.
	\]
	So if all test images are predicted correctly then $GAP_{W\text{ }}=0$ and the
	idea is to exclude the worst window.

%	Note that if test image $B$ is in class $B_{test}^{i}$ and is not
%	predicted correctly by WNN$_{W}$ algorithm then%
%	\[
%	Dist_{W}(B,A_{train}^{i})>\min_{j\in\left\{1,...,10\right\}}%
%	Dist_{W}(B,A_{train}^{j}).
%	\]
%	We iteratively apply the above algorithm to the remaining windows until the desired number of excluded windows is obtained.
	
	%We then exclude the next window from the remaining 783 windows according to the above procedure and repeat the procedure until the desired number of windows remain.
	
\begin{figure}[h!]
	\begin{center}
		\includegraphics[
		height=2.7224in,
		width=3.6253in
		]%
		{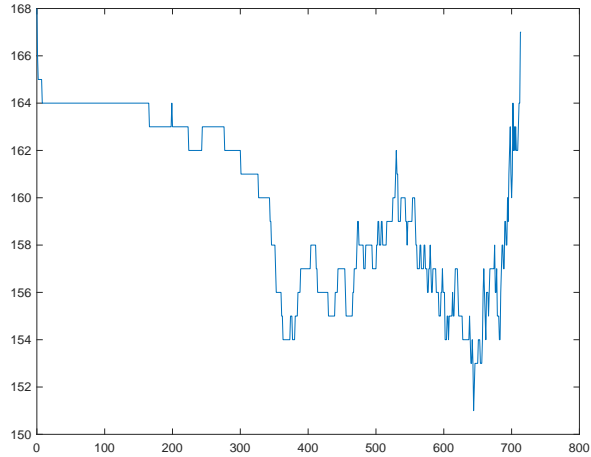}%
	\end{center}
	\caption{Number of errors versus number of excluded windows.}
	\label{fig:error_excl_win_1}
\end{figure}

	Let us next consider an example on EMNIST. We apply the above algorithm to Set 4 (recall that it consists of 30$\times10^{6}$ images). Then the number of errors with respect to the number of excluded windows $K$ can be seen in Figure \ref{fig:error_excl_win_1}.%
	%TCIMACRO{\FRAME{ftbpF}{3.6253in}{2.7224in}{0in}{}{}{mm_40000.eps}%
	%width 3.6253in;  height 2.7224in;  depth 0in;  original-width 7.7798in;
	%original-height 5.8271in;  cropleft "0";  croptop "1";  cropright "1";
	%cropbottom "0";  filename 'MM_40000.eps';file-properties "XNPEU";}}}%
	%BeginExpansion
In particular when 100 windows are used, i.e. $K=684$, the number of errors
	will be 156. Using only 60 windows the number of errors increase to 167 which still is less than 168 which was obtained using all 784 windows. 
	
\begin{figure}[h!]
		\begin{center}
			\includegraphics[
			height=2.7224in,
			width=3.6253in
			]%
			{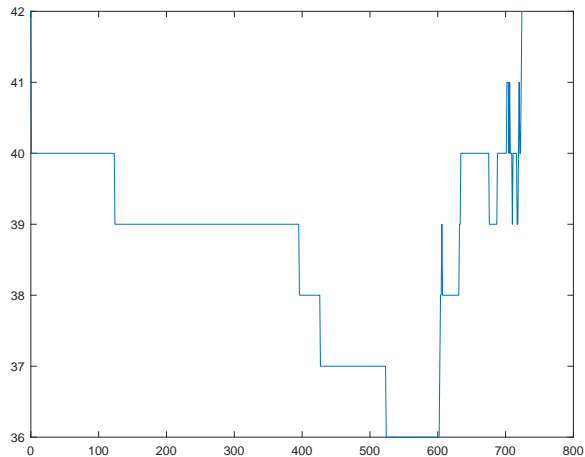}%
		\end{center}
		\caption{Number of errors versus number of excluded windows using validation set.}
		\label{fig:error_excl_win_2}
	\end{figure}

	However, note that constructing the set of excluded windows by using the whole test set is not correct. We therefore divide the test set of 40000 images randomly into two subsets: the validation set of 30000 images (3000 images for each digit) will be used for determining the number of excluded windows and the remaining set of 10000 images will be the new test set. The resulting graph of errors can be seen in Figure \ref{fig:error_excl_win_2}.%
	%TCIMACRO{\FRAME{ftbpF}{3.6253in}{2.7224in}{0in}{}{}{mm_10000.eps}%
	%width 3.6253in;  height 2.7224in;  depth 0in;  original-width 7.7798in;
	%original-height 5.8271in;  cropleft "0";  croptop "1";  cropright "1";
	%cropbottom "0";  filename 'MM_10000.eps';file-properties "XNPEU";}}}%
	%BeginExpansion
In particular, if we use just 50 windows then the number of errors will be 42 corresponding to an error rate of 0.42\%. Recall that this error rate is the same as when using all 784 windows on the original test set.

\section*{Acknowledgements}
We acknowledge computational resources from the National Supercomputer Centre at Linköping University through the project LiU-compute-2021-41: Nearest Neighbour Classifier.

\bibliographystyle{plain} % We choose the "plain" reference style
\bibliography{ref} % Entries are in the refs.bib file	

\end{document}